# SEGMENTAÇÃO E CONTAGEM DE TRONCOS DE MADEIRA UTILIZANDO DEEP LEARNING E PROCESSAMENTO DE IMAGENS

**João V. C. Mazzochin[1], Gustavo Tiecker[2], Érick Oliveira Rodrigues[3]**

[1]Departamento Acadêmico de Informática (DAINF) - Universidade Tecnológica Federal do Paraná (UTFPR), Pato Branco - PR - Brazil.

{joaomazzochin, gustavotiecker}@alunos.utfpr.edu.br,
erickrodrigues@utfpr.edu.br

***Abstract.*** *Counting objects in images is a pattern recognition problem that focuses on identifying an element to determine its incidence and is approached in the literature as Visual Object Counting (VOC). In this work, we propose a methodology to count wood logs. First, wood logs are segmented from the image background. This first segmentation step is obtained using the Pix2Pix framework that implements Conditional Generative Adversarial Networks (CGANs). Second, the clusters are counted using Connected Components. The average accuracy of the segmentation exceeds 89% while the average amount of wood logs identified based on total accounted is over 97%.*

***Resumo.*** *A contagem de objetos em uma imagem é um problema de reconhecimento de padrões que visa identificar um elemento e determinar sua incidência e é bordada na literatura como Visual Object Counting (VOC). No presente trabalho propomos uma metodologia para contabilizar troncos de madeira. Primeiro os troncos são segmentados. A primeira etapa de segmentação é obtida usando o framework Pix2Pix que implementa Redes Adversariais Generativas Condicionais (CGANs). Em seguida, a contagem é realizada utilizando Connected Components. A média de acurácia na segmentação excede 89% enquanto a média de troncos contabilizados em relação ao total observado ultrapassa 97%.*

## 1. Introdução

A contagem de objetos é uma atividade importante e recorrente em diversos nichos, como industrial e laboratorial. Nesses casos o processo de contagem pode ser determinante, influenciando diretamente no resultado final de uma análise detalhada.

Em um contexto industrial organizacional, em especial as produtoras de artefatos físicos em geral, a contagem de estoque pode vir a ser determinante na gestão de recursos. Junior, Walker e Santos (2020) inferem que a contagem de itens em estoque tem influência direta no faturamento e eficiência de uma organização.

Tanto no âmbito industrial como laboratorial, muitas vezes a técnica manual é a mais utilizada, abrindo campo para pesquisa nestas áreas visando o desenvolvimento de novas soluções com finalidade de agilizar e tornar os diagnósticos mais precisos no contexto laboratorial e melhorar o gerenciamento de recursos no âmbito industrial organizacional.

No presente trabalho foi desenvolvido uma metodologia voltada ao contexto industrial, mais precisamente focada no reconhecimento de troncos de madeiras, segmentação e a realização da contagem de incidências destes em uma imagem. Para tal, foi aplicado a rede neural Pix2Pix, variando hiper parâmetros e aplicando pré-processamentos de imagem, mais precisamente operações morfológicas. Complementar a isso, a contagem de incidências dos troncos foi realizada com um algoritmo *Connected Components* partindo da segmentação gerada pelo modelo Pix2Pix.

## 2. Referencial Teórico

As *Generative Adversarial Networks* (GANs) foram introduzidas como uma estrutura alternativa para o treinamento de modelos generativos, visando contornar a dificuldade de aproximar muitos cálculos probabilísticos intratáveis. Entretanto, em um modelo gerador não condicionado, não há controle sobre os modos em que os dados estão sendo gerados [Mirza e Osindero 2014].

As GANs podem ser estendidas a um modelo condicional *Conditional Generative Adversarial Networks* (CGANs), onde o gerador e o discriminador são condicionados a informações extras *y*. Ao condicionar o modelo à informações adicionais, é possível direcionar o processo de geração de dados, onde *y* pode ser qualquer tipo de informação auxiliar, como rótulos de classe ou dados de outra modalidade. É possível utilizar a condição como uma camada de entrada adicional tanto no modelo gerador, quanto no modelo discriminador [Mirza e Osindero 2014].

$$\min_G \max_D V(D, G) = \mathbb{E}_{x \sim p_{data}(x)}[log D(x|y)] + \mathbb{E}_{z \sim p_z(z)}[log(1 - D(G(z|y)))].$$

**Equação 1. Formula Conditional Generative Adversarial Networks**

Em resumo, o discriminador e gerador jogam o seguinte jogo *minimax* (em teoria da decisão é um método para minimizar a possível perda máxima) de dois jogadores com função de valor *V(D,G)*.

Segundo Isola et al. (2017), essas Redes Neurais não apenas aprendem a mapear uma imagem de entrada para uma imagem de saída, mas também aprendem uma função de perda para treinar e aperfeiçoar esse mapeamento. A tarefa de transformar uma possível representação de uma imagem em outra baseada em um conjunto suficiente de dados é conhecida como transformação *image-to-image*, essa técnica se resume na predição de *pixels* a partir de *pixels*. O Pix2Pix, framework utilizado para segmentação, implementa o uso de CGANS e por consequência apresenta uma solução comum para tarefas que exigem esse tipo de técnica.

As segmentações geradas pelo Pix2Pix são submetidas a um algoritmo de contagem chamado *Connected Components* o qual consiste em rotular componentes conectados em uma imagem binária. Ao utilizar a operação de atribuição de rótulos, é possível transformar uma imagem binária em uma imagem simbólica, onde todos os *pixels* pertencentes ao mesmo componente conectado recebem o mesmo rótulo [HE et al. 2009]. Esta abordagem é utilizada em nosso algoritmo de contagem de troncos, visando identificar grupos de componentes conectados que podem ser classificados como instâncias de um mesmo tronco a ser contabilizado.

Durante a pesquisa não foram encontrados registros da utilização do Pix2Pix para a identificação de objetos em imagens. Além disso, poucos trabalhos possuem um contexto semelhante lidando com a identificação e contagem de troncos, dentre estes apenas o de Yella e Dougherty (2013) possuem dados suficientes para uma comparação.

Yella e Dougherty (2013) fazem a segmentação de imagem baseada em cor utilizando K-Means, sua base de dados possui a característica onde os troncos de madeira destacam-se em cor em relação ao restante das partes da imagem. Logo esse detalhe foi utilizado como uma característica na segmentação da imagem, separando a imagem em regiões baseando-se na homogeneidade e heterogeneidade da mesma. Na sequência, para fazer a identificação de troncos foi feita a utilização do algoritmo de clusterização Circular Hough Transform, o qual consiste em uma técnica de extração utilizada para identificar círculos em imagens. A Figura 1 apresenta o processo de segmentação e contagem empregado por Yella e Dougherty (2013).

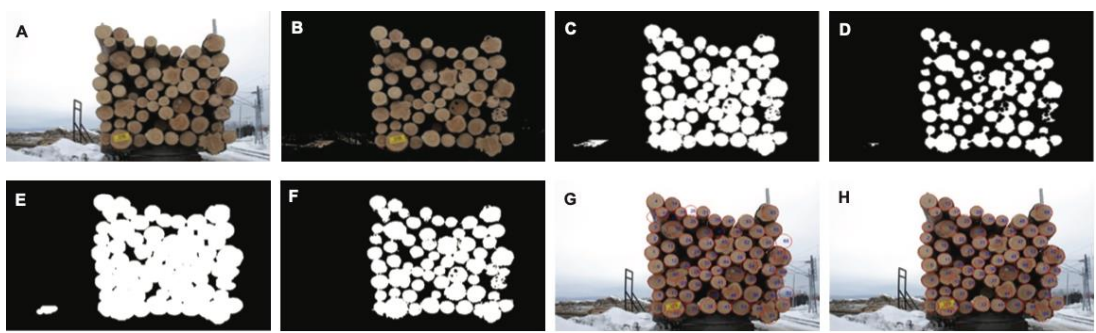

**Figura 1. Imagem retirada do artigo de Yella e Dougherty (2013) apresentando o processo de segmentação e contagem por eles empregado**

Yella e Dougherty (2013) realizaram testes que diferem dos aplicados no presente trabalho, eles envolveram dois observadores para realizar a contagem e formar a análise subjetiva, realizando a comparação destas duas análises com o número de troncos detectados pelo algoritmo de visão de máquina.

## 5. Método

As imagens da base de dados são compostas por agrupamentos de faces de madeiras que compõem uma pilha, mais especificamente faces de troncos de eucalipto. A base de dados é composta por 132 imagens que foram capturadas dentro do intervalo de 8 horas às 17 horas em cidades do sudoeste do Paraná. As imagens capturadas apresentam principalmente variação na iluminação, angulação, distância do objeto a ser identificado e adversidades (folhas, galhos, cascas de árvores, etc).

O método do presente trabalho visa a identificação de incidências de troncos em uma imagem, tendo como base duas partes principais: segmentação e contagem. O Pix2Pix foi utilizado na segmentação pelo fato de realizá-la *pixel* a *pixel*. Enquanto para contagem foi utilizado o algoritmo *Connected Components*.

Para o treinamento do modelo de segmentação, primeiramente uma imagem real é segmentada manualmente. Na sequência, a imagem original e a imagem segmentada são submetidas ao Pix2Pix.

Segundo Zhang (1996), para avaliar a qualidade da segmentação existem duas formas principais: avaliação subjetiva e avaliação supervisionada. A metodologia de avaliação empregada foi a supervisionada. Para a compreensão desta, é fundamental o entendimento de *Ground Truth*, TP (*True Positive*), FP (*False Positive*), TN (*True Negative*) e FN (*False Negative*). O conceito clássico de algoritmos de avaliação é baseado principalmente em TP, FP, TN e FN [Taha e Hanbury 2015].

Para os testes de performance do modelo, foram utilizadas as medidas *Accuracy*, *F1 Score*, *Kappa* e *Intersection Over Union* (IoU) [Hu et al. 2020]. Estes testes foram realizados na intenção de identificar qual é o número de amostras necessárias para o modelo gerar resultados satisfatórios. A Figura 2 apresenta um gráfico onde é expressa a média de cada uma das medidas para iterações variando o número de amostras na base de treinamento.

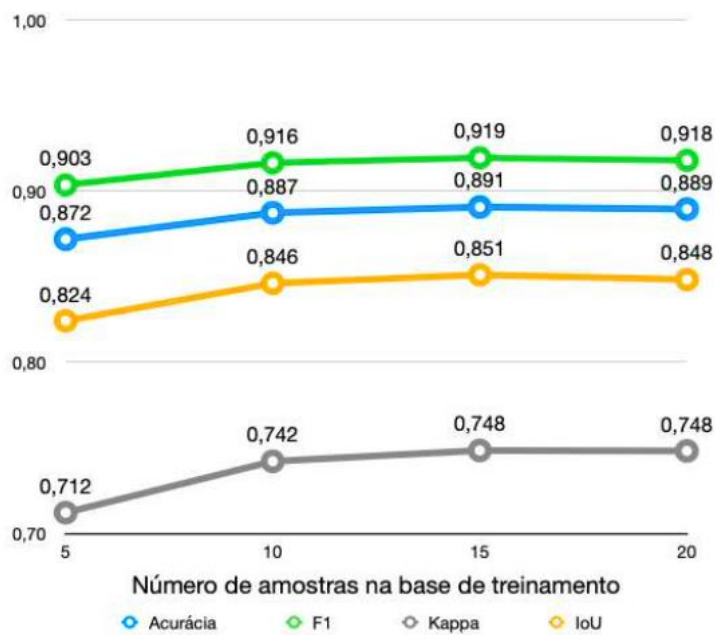

**Figura 2. Média aritmética dos índices calculados**

Note que, a queda gerada entre 15 e 20 amostras na base de treino pode se tratar de um mínimo local e não o mínimo global. Por conta disso, não foi o único fator determinante para a interrupção do processo de segmentação. Dentre outros fatores, o aumento do número de imagens segmentadas também não trouxe grandes ganhos de performance. Logo, definimos 15 amostras como ideal para a composição da base de treino na realização dos testes efetuados neste trabalho. Apesar do baixo número de amostras, é necessário avaliar a quantidade de *pixels* na resolução da imagem aceita como entrada pelo Pix2Pix (256x256), multiplicando pelo número de imagens utilizadas para treino do modelo, temos um total de 983040 *pixels* utilizados como entrada, o que é um número considerável de instâncias para treinamento.

A Figura 3 apresenta um fluxograma do processo realizado, desde a etapa de treinamento do modelo até avaliação da qualidade da segmentação.

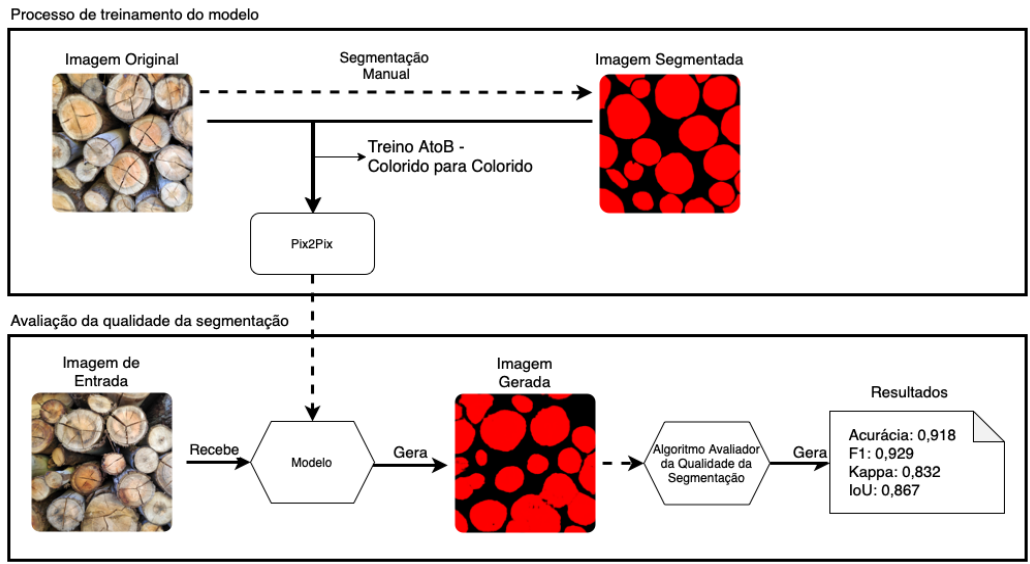

**Figura 3. Fluxograma do resultado da segmentação**

Para entender nosso método de contagem, primeiramente é necessário compreender os problemas da contagem. Assim como na segmentação, temos uma imagem de entrada representada por uma matriz. O primeiro desafio é marcar na imagem os *pixels* que fazem parte da face do tronco, gerando uma imagem binária que será a entrada para os algoritmos que buscam identificar o número de troncos existentes. Para realizar a contagem é necessário identificar os grupos de *pixels* denominados *clusters*, e é nesta etapa que o principal problema ocorre. Diferentes *clusters* podem ser conectados por interseções que interferem diretamente na identificação dos mesmos. Somado a isso, os ruídos gerados pelo modelo, bem como falhas de segmentação que transpassam as extremidades de uma face podem ser reconhecidos como *clusters* distintos, os quais, erroneamente, também são somados ao montante da contagem. A Figura 4 apresenta um exemplo claro de ruídos e falhas de segmentação.

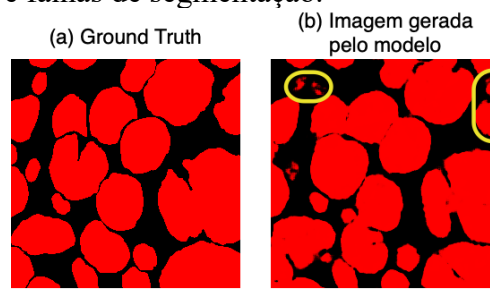

**Figura 4. Exemplo de ruído e falha na segmentação**

Para contornar estes problemas foi treinado um novo modelo, este voltado somente para segmentar imagens com o mínimo de interseção e ruídos. Para tal, utilizamos da mesma metodologia aplicada à segmentação, porém com uma etapa extra. Antes de treinar o modelo, foi aplicado às amostras de expectativa da base de treino a operação morfológica de erosão, transformando do *ground truth,* deixando evidente que não deve ocorrer a interseção entre os *clusters*. A Figura 5 apresenta: a segmentação manual, a comparação entre a geração do modelo sem aplicar nenhum pré-processamento à base de treino, a geração do modelo aplicando o pré-processamento de erosão à base de treino e a subtração de ambas as gerações.

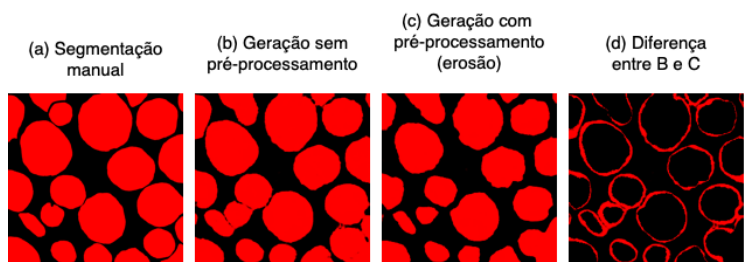

**Figura 5. Comparativo entre imagens geradas pelo modelo com e sem pré-processamento aplicado à base de treinamento**

Note que, as duas gerações ainda apresentam interseções comparado a segmentação *ground truth* presente na Figura 5a, mas claramente na Figura 5c houve uma melhora na geração do modelo. Por esse motivo, quando mencionada a base de treino nos testes de contagem seguintes, trata-se da base já erodida (operação da morfologia matemática). A Figura 6 apresenta um fluxograma do processo realizado para a obtenção dos resultados da contagem.

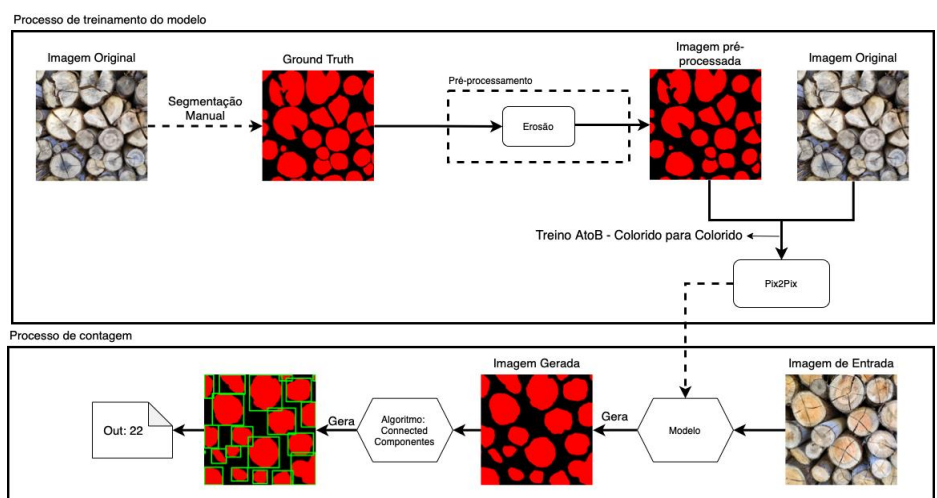

**Figura 6. Fluxograma do resultado da contagem**

Visando avaliar a performance do método de contagem empregado, realizamos a análise subjetiva das saídas geradas pelo modelo. Por meio desta, é possível observar de forma simples o que se espera da tarefa de contagem, avaliando quantos troncos foram classificados de maneira correta. O objetivo principal é encontrar interseções ou troncos que não existem na imagem original. Com esse dado é possível avaliar o desempenho dos algoritmos, verificando se as instâncias estão sendo contabilizadas de forma honesta.

## 6. Resultados

Após treinado o modelo, 10 novas imagens foram submetidas para a coleta de resultados. O modelo treinado para segmentação gerou a segmentação equivalente de cada entrada que foram posteriormente submetidas ao algoritmo avaliador que calculou os índices *Accuracy*, *F1 Score*, *Kappa* e *IoU* para cada uma das imagens (Figura 7).

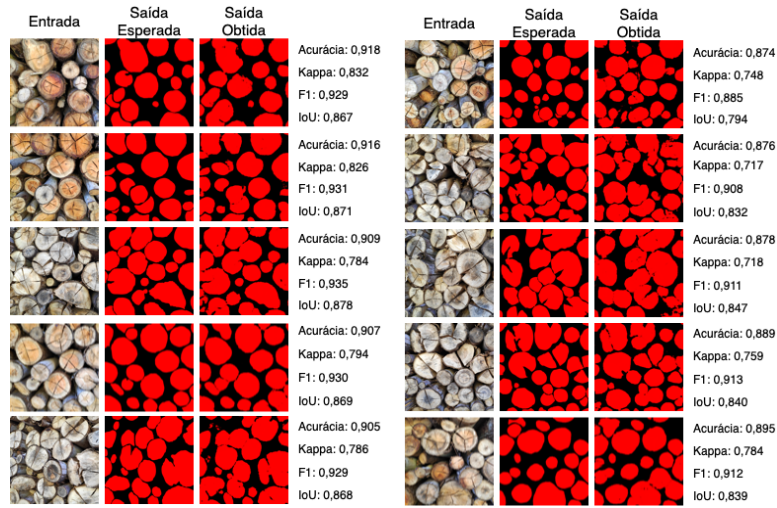

**Figura 7. Resultados da segmentação**

Para coleta de resultados da contagem foi treinado um novo modelo. As imagens utilizadas para realizar o treinamento deste modelo foram segmentadas manualmente. Posteriormente, foi aplicada a operação morfológica de erosão. Com isso, obtivemos uma base de treino que consiste em amostras pré-processadas. Durante a etapa da pesquisa sobre a contagem, foram realizados testes com outros algoritmos como o *Circular Hough Transform* e Morphological Reconstruction. No entanto, o *Connected Components* obteve melhores resultados. A Listagem 1 apresenta o pseudo algoritmo de *Connected Components* utilizado no presente trabalho:

```
1:   se o pixel (x, y) tiver "0" então
2:       Não fazer nada e continuar para o próximo pixel (x+1, y)
3:   senão se o pixel (x-1, y-1) tiver rotulado então
4:       Rotule também o pixel (x, y)
5:   senão se nenhum dos pixels (x-1, y) OR (x, y-1) estiverem rotulados então
6:       Incremente o número de rótulos numerados e rotule o ultimo pixel (x, y) identificado
7:   senão se os pixels (x-1, y) XOR (x, y-1) estiverem rotulados então
8:       Rotule igualmente o pixel (x, y)
9:   senão se ambos os pixel (x-1, y) AND (x, y-1) estiverem rotulados então
10:      Rotule o pixel (x, y) igualmente ao pixel (x-1, y)
11:      Registre a equivalência se os rótulos dos pixels (x-1, y) e (x, y-1) não forem iguais
12:  fim se
```

**Listagem 1. Listagem do pseudo algoritmo de Connected Components**

Primeiramente lemos uma imagem e convertemos ela para escala de cinza, na sequência extraímos da função *connectComponentsWithStats()*, implementada pela biblioteca *OpenCV*, a quantidade de rótulos contabilizados, bem como a variável *stats* que consiste em uma matriz de estatísticas calculada pela função. Posteriormente, esta é usada dentro de um loop no número de rótulos identificados para gerar um retângulo contornando cada um dos *clusters* identificados. Por fim, imprimimos o número de rótulos identificados e salvamos a imagem gerada.

Assim como nos resultados da segmentação, após treinado o modelo, 10 novas imagens foram submetidas para a coleta de resultados, o modelo gerou a segmentação correspondente para cada uma, que foram posteriormente submetidas ao algorítimo

*Connected Components* que calculou quantos grupos de *clusters* existem em cada uma das imagens. A Figura 8 apresenta os resultados obtidos na etapa de contagem.

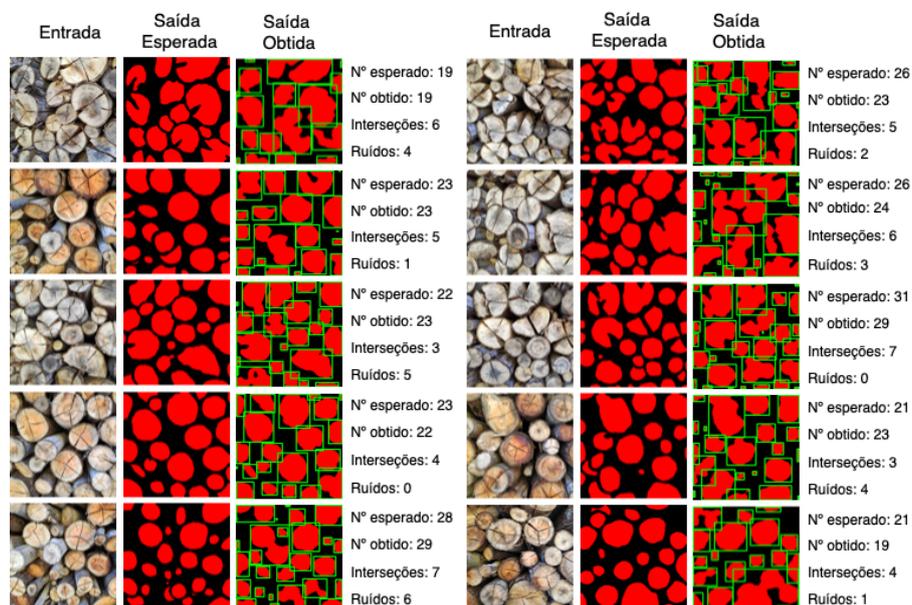

**Figura 8. Resultados da contagem**

Quando comparados nossos resultados da contagem com os obtidos por Yella e Dougherty (2013) é importante notar que a comparação é em partes injusta, visto que, não tivemos possibilidade de realizar testes com as mesmas amostras. Além disso, somente com a descrição realizada pelos autores não foi possível realizar a implementação da metodologia por eles empregada. Nos resultados apresentados por Yella e Dougherty (2013) não constam métricas relevantes para a comparação, como por exemplo, a quantidade de troncos que foram classificados de forma errada, por serem inexistentes, bem como troncos que deixaram de ser classificados. Com as informações contidas conseguimos extrair apenas a acurácia total de seu método em relação ao total de troncos observados. A Figura 9 apresenta a comparação entre os trabalhos.

| Abordagem | % média de troncos contabilizados em relação ao total observado |
|---|---|
| Mazzochin e Tiecker | 97,50 |
| Yella e Dougherty | 79,73 |

**Figura 9. Comparação dos resultados de contagem do presente trabalho e da abordagem aplicada por Yella e Dougherty (2013)**

Vale reforçar que as médias obtidas são provenientes de bases de dados diferentes, cada uma com adversidades e particularidades, além disso, os dados comparados não foram coletados de forma equivalente. Logo, a comparação não pode ser considerada justa do ponto de vista científico.

## 7. Conclusão

O presente trabalho introduz uma metodologia de segmentação e contagem de troncos utilizando o framework Pix2Pix e técnicas de processamento de imagem, no intuito de gerar um modelo capaz de segmentar as incidências do objeto a ser identificado. O método proposto consiste na segmentação manual de imagens de troncos de madeira, que juntamente com a imagem original, são utilizadas para treinamento de um modelo que deve ser capaz de receber uma nova entrada e gerar a segmentação equivalente. As segmentações manuais foram submetidas ao pré-processamento de erosão antes de serem submetidas ao treinamento de um novo modelo. Posteriormente, as imagens geradas por esse novo modelo são submetidas a um algoritmo que visa identificar grupos de componentes que estejam conectados, realizando uma clusterização por densidade. Nossa abordagem se distingue pelo uso do Pix2Pix, junto com a abordagem de Connected Components, visando a execução de uma tarefa de contagem, algo não abordado na literatura.

A partir dos resultados obtidos, é possível concluir que embora o Pix2Pix tenha uma ótima performance em aprendizado para segmentação, o mesmo pode levar a problemas na fase da contagem devido a geração de interseções e ruídos. Apesar desses problemas, durante o processo de pesquisa notamos que aprofundando o estudo de morfologia matemática, com técnicas diferentes de pré-processamento e pós-processamento é possível tornar a abordagem mais precisa no momento da contagem.

Por fim, no contexto da segmentação, nosso trabalho possui uma acurácia superior a 89% e uma taxa de verdadeiro positivo por *pixel* alta. Embora a contagem ainda precise de melhorias, se não forem consideradas as interseções e os ruídos, a média da acurácia na contabilização de troncos em relação ao total observado ultrapassa 97%.

Como consideração adicional, o estado atual da abordagem de segmentação também já pode ser usado para estimar o volume de uma pilha de troncos de forma satisfatória.

## 7. Referências